\documentclass[accepted]{uai2021} 

\usepackage[american]{babel}

\usepackage{natbib} 
\usepackage{mathtools} 
\usepackage{siunitx} 
\usepackage{booktabs} 
\usepackage{tikz} 

\usepackage{commands} 
\DeclareMathAlphabet{\mathcal}{OMS}{cmsy}{m}{n}

\title{A Bayesian Approach to Learning Bandit Structure in Markov Decision Processes}

\author[1]{\href{mailto:Kelly W. Zhang <kellywzhang@seas.harvard.edu>?Subject=A Bayesian Approach to Learning Bandit Structure in Markov Decision Processes}{Kelly W. Zhang}{}} 
\author[2]{Omer Gottesman}
\author[1]{Finale Doshi-Velez}
\affil[1]{%
    Department of Computer Science, Harvard University
}
\affil[2]{%
   Department of Computer Science, Brown University
}

\begin{document}
\maketitle

\begin{abstract}
	In the reinforcement learning literature, there are many algorithms developed for either Contextual Bandit (CB) or Markov Decision Processes (MDP) environments. However, when deploying reinforcement learning algorithms in the real world, even with domain expertise, it is often difficult to know whether it is appropriate to treat a sequential decision making problem as a CB or an MDP. In other words, do actions affect future states, or only the immediate rewards? Making the wrong assumption regarding the nature of the environment can lead to inefficient learning, or even prevent the algorithm from ever learning an optimal policy, even with infinite data. In this work we develop an online algorithm that uses a Bayesian hypothesis testing approach to learn the nature of the environment. Our algorithm allows practitioners to incorporate prior knowledge about whether the environment is that of a CB or an MDP, and effectively interpolate between classical CB and MDP-based algorithms to mitigate against the effects of misspecifying the environment. We perform simulations and demonstrate that in CB settings our algorithm achieves lower regret than MDP-based algorithms, while in non-bandit MDP settings our algorithm is able to learn the optimal policy, often achieving comparable regret to MDP-based algorithms.
\end{abstract}



\section{Introduction}

\begin{figure*}
		\centerline{
		\includegraphics[width=0.37\linewidth]{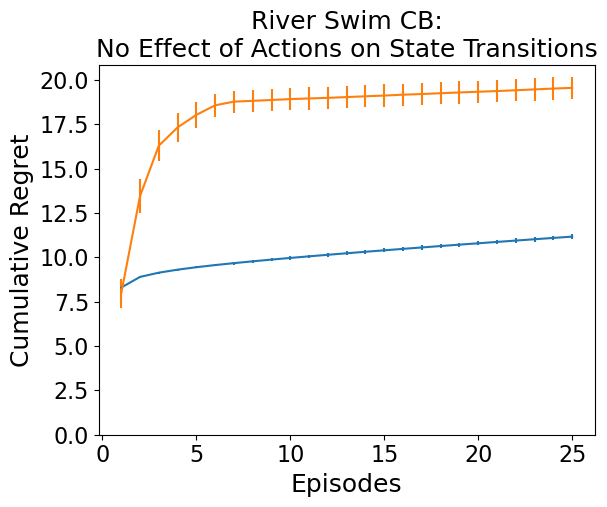}
  		~~~~~~~~
  		\includegraphics[width=0.38\linewidth]{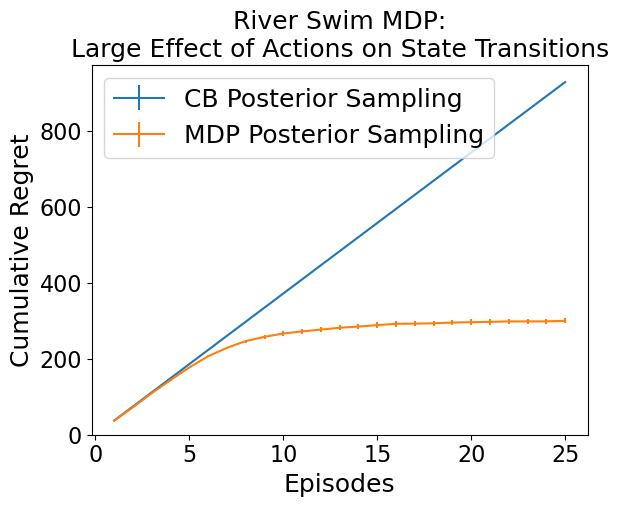}
  		}
  		
  		\caption{Above we plot the cumulative regret of a contextual bandit posterior sampling algorithm and a MDP posterior sampling algorithm in two environments (finite horizon 100, 6 states). The MDP environment is the river swim environment (see Figure \ref{fig:riverswimenv}).
  		\bo{The contextual bandit environment that is identical to the MDP environment, except the state transition probabilities are uniform over all states.} The error bars are standard errors for the estimates (100 repetitions).}
  		\label{fig:cb_vs_mdp}
	\end{figure*}

	Sequential decision making problems are commonly analyzed using two different frameworks: Contextual Bandits (CB) and Markov Decision Processes (MDP). In contextual bandit environments, it is assumed that states evolve independently of actions selected by the algorithm, whereas for MDP environments, action selections affect state transition probabilities.\footnote{Note that CB environments is simply a special case of MDPs.} 
	As a consequence in MDP environments, action selections may have long term effects on which states are visited in the future; therefore an algorithm designed for a bandit environment, which does not take such long term effects into account, may never be able to learn an optimal policy. 
	On the other hand, if an environment is known satisfy CB assumptions, this restriction of the problem can be leveraged for more efficient learning, and therefore algorithms designed for bandit environments learn more effectively, as long as the CB assumptions hold. 
	To illustrate this, in Figure \ref{fig:cb_vs_mdp} we compare the performance of CB and MDP posterior sampling algorithms in both a contextual bandit and an MDP setting, and demonstrate that each algorithm significantly outperforms its competitor in the setting it was designed for.
	
    Reinforcement learning (RL) algorithms are typically developed assuming the environment falls strictly within one of these two frameworks. However, when deploying RL algorithms in the real world, it is often unknown which one of these frameworks one should assume. 
	For example, consider a mobile health application aimed to help users increase their step count \citep{liao2020personalized}. A few times a day, the app's algorithm uses the user's state information (e.g. recent app engagement, local weather, time of day, etc.) to decide whether or not to send the user a message encouraging them to take a walk.
	It is not immediately clear whether to use a contextual bandit or an MDP algorithm in this problem setting. It may be that sending notification strongly affects the future state of the user---for example, notifications that annoy the user could lead them to disengage; 
	if this is a possibility, an MDP algorithm is favored, because a CB algorithm will not be able to learn the optimal policy.
	On the other hand, it may be that users' states are not greatly affected by notification---for example, users' responsiveness may depend purely on factors such as weather or their work schedule, and past messages do not play a role. In this setting, a CB algorithm could perform better, and making a general MDP assumption can lead to inefficient learning and incurring greater regret.
	
	In this work, we develop the \textit{Bayesian Hypothesis Testing  Reinforcement Learning (BHT-RL)} algorithm, which is an online algorithm for the finite-horizon episodic setting, which utilizes Bayesian Hypothesis Testing to learn whether the environment is that of a CB or a classical MDP.
	Practitioners specify the prior probability on the environment being a bandit, as well as choose two RL algorithms: one CB algorithm and one MDP algorithm. At the start of each episode, BHT-RL selects which algorithm to use, CB vs. MDP, according to the posterior probability of the environment being a bandit. The choice of prior probability allows the practitioner to choose the amount to ``regularize'' the MDP algorithm through the amount of evidence needed to favor using the MDP algorithm over the CB algorithm.
	We empirically demonstrate that in terms of regret minimization, BHT-RL outperforms CB and MDP algorithms in their respective misspecified environments, and that BHT-RL is often competitive with both CB and MDP algorithms each in the environments they were designed for. Moreover, BHT-RL results in significantly better regret minimization empirically in both contextual bandit and MDP environments compared prior work on learning bandit structure in MDPs \citep{zanette2018structure, zanette2019tighter}. 
	Additionally, when posterior sampling algorithms are chosen for the contextual bandit and MDP algorithms, then the BHT-RL algorithm can be interpreted as posteroir sampling (PS) with an additional prior that can up-weight the prior probability that environment is a bandit; thus, the Bayesian regret bounds for standard MDP-PS apply \citep{osband2013psrl}.
	

	

\section{Related Work}

An open problem in RL theory is understanding how sample complexity depends on the planning horizon (in infinite horizon problems, this is the discount factor) \citep{jiang2018open}.
\citet{jiang2015dependence} showed that when learning the optimal policy from data in MDP problems, longer planning horizons increase the size of the set of policies one searches over.
As a result, often when one has small amounts of data it is better to use a smaller planning horizon than the evaluation horizon as a method of regularization to prevent overfitting to the data \citep{arumugam2018mitigating}.
However, using a shorter planning horizon can also prevent algorithms from ever learning the optimal policy---for example, a CB algorithm, which has a planning horizon of $1$, will never be able to learn the optimal policy in most MDP environments. Since the BHT-RL algorithm interpolates between CB and MDP algorithms, it can be viewed as a regularized version of standard MDP algorithms which reduces the planning horizon, i.e., using a CB algorithm, when the evidence that the environment is an MDP is low.


There have been several works developing low regret algorithms for both CB and MDP environments.
\citet{jiang2017cdp} develop the OLIVE algorithm and prove bounds for it in a variety of sequential decision making problems when the the Bellman rank is known. However, since the Bellman rank of problems is generally unknown in real world problems, OLIVE cannot be used in practice. 
\citet{zanette2018structure} develop the UBEV-S RL algorithm, which they prove has near optimal regret in the MDP setting and has regret that scales better than that of OLIVE in the CB setting \citep{zanette2018structure}.
Later, \citet{zanette2019tighter} developed the EULER algorithm, which improves upon UBEV-S, and has optimal regret bounds in both the contextual bandit and MDP settings \citep{zanette2019tighter}.
Both UBEV-S and EULER are upper confidence bound based methods that construct confidence bounds for the next timestep reward and future value, and then execute the most optimistic policy within those bounds. Although there have only been Bayesian regret bounds (in contrast to frequentist regret bounds) proven for posterior sampling on MDPs \citep{osband2013psrl}, it has been shown in previous work that posterior sampling RL algorithms generally outperform confidence bound based algorithms empirically \citep{osband2013psrl, osband2017posterior, russo2014learning}, and in our experiment we see that BHT-RL outperforms both UBEV-S and EULER.



For BHT-RL, we pool the state transition counts for different actions in the same state together. 
\citet{asmuth2009bayesian} call this approach the tied Dirichlet model. However, they also assume that the experimenter has apriori knowledge and chooses before the study is run whether to assume the tied or regular Dirichlet model on the transition probabilities.
In contrast, we will aim to \textit{learn} whether in each state it is better to use the tied Dirichlet model or the standard one.
To do this we will use Bayesian hypothesis testing \citep{berger2012bht}.
Bayesian hypothesis testing is related to Bayesian model selection because the posterior probabilities of the null versus the alternative models are a function of the Bayes factor, which is used in model selection to compare the relative plausibilities of two different models or hypotheses.

\section{Bayesian Hypothesis Testing RL}

\subsection{Problem Setting}

We define random variables for the states $S_t \in \MC{S}$, random variables for action selections $A_t \in \MC{A}$, and rewards $R_t \in \real$. 
We also define $\theta$, which parameterizes the environment, i.e., given $\theta$ we know the expected rewards $E_\theta [ R_t | A_t = a, S_t = s ]$ and the transition probabilities $P_\theta ( S_{t+1} = s' | S_t = s, A_t = a )$.
We assume a finite-horizon episodic setting, so the data collected is made up of episodes each of length $H$.
For example, for the $k^{th}$ episode, we have the data $( A_{t_k + h}, S_{t_k + h}, R_{t_k + h} )_{h=1}^H$, where $t_k := k H$.
We define $\MC{H}_{t_k} = \{ ( S_{t_k + h}, A_{t_k + h}, R_{t_k + h} )_{h=1}^H \}_{k'=0}^{k-1} = \{ ( A_t, S_t, R_t ) \}_{t=1}^{t_k}$ to be history at time $t_k$. 
Note that we define our policies to be $\pi_k$ to be $\sigma( \MC{H}_{t_k} )$-measurable functions from $\MC{S} \by [1 \colon H]$ to $|\MC{A}|$-dimensional simplex.
So, our actions $A_{t_k + h} \sim \pi_k( S_{t_k+h}, h)$ are chosen according to the policy. 
Note that the policy takes the time-step in the episode, $h$, as an input because in the finite horizon setting the optimal policy can change depending on the timestep in the episode.

\subsection{Algorithm Definition}


For our Bayesian Hypothesis Testing method we define the following null and alternative hypotheses. Throughout, we focus on the discrete state setting, but these hypotheses and the BHT-RL method could be generalized to continuous states, when one has a model for the transition probabilities.

\paragraph{Null hypothesis $H_0$:} 
\textit{Action selections do not affect transition probabilities, i.e., for all $a \in \MC{A}$, $s, s' \in \MC{S}$,} 
\begin{equation*}
    P_\theta( S_{t+1} = s' | S_t = s, A_t = a ) = P_\theta( S_{t+1} = s' | S_t = s )
\end{equation*}
Under the null hypothesis we model our data as generated by the following process:
\begin{itemize}
	\item For each $s \in \MC{S}$ we draw $\tilde{\theta}_s$ from a prior distribution over the transitions. For example, we will use $\tilde{\theta}_s \sim \TN{Dirichlet}( \bs{\alpha} )$ for some $\MC{S}$-dimensional vector $\bs{\alpha}$ with positive entries in our derivations and simulations.
	\item For all $t \in [1 \colon T]$ such that $S_t = s$, we have that $S_{t+1} \sim \TN{Categorical}( \tilde{\theta}_s )$.
\end{itemize}
		
\paragraph{Alternative hypothesis $H_1$:} 
\textit{Action selections do affect transition probabilities, i.e., for some $a, a' \in \MC{A}$, $s, s' \in \MC{S}$,} 
\begin{equation*}
    P_\theta( S_{t+1} = s' | S_t = s, A_t = a ) \not= P_\theta( S_{t+1} = s' | S_t = s, A_t = a' )
\end{equation*} 
Under the alternative hypothesis we model our data as generated by the following process:
\begin{itemize}
	\item For each $s \in \MC{S}$ and each $a \in \MC{A}$ we draw $\tilde{\theta}_{s,a}$ from a prior distribution over the transition probabilities. For example, $\tilde{\theta}_{s,a} \sim \TN{Dirichlet}( \bs{\alpha} )$.
	\item For all $t \in [1 \colon T]$ such that $S_t = s$ and $A_t = a$, we have that $S_{t+1} \sim \TN{Categorical}( \bs{\theta}_{s,a} )$.
\end{itemize}

BHT-RL, as defined in Algorithm \ref{alg:bhtrl}, requires one to choose a contextual bandit algorithm (denoted $\pi^{\TN{CB}}$), an MDP based algorithm (denoted $\pi^{\TN{MDP}}$), a generative model for the transition probabilities under both hypotheses, and a prior probability over the hypotheses $P(H_0)$ and $P(H_1) = 1 - P(H_0)$. At the start of each episode, BHT-RL selects which algorithm to use, $\pi^{\TN{CB}}$ vs. $\pi^{\TN{MDP}}$, according to the posterior probability of the null hypothesis.
Note that if we set the prior probability of the null $P(H_0)$ to $1$ the BHT-RL algorithm is equivalent to  $\pi^{\TN{CB}}$ and when setting $P(H_0)$ to $0$ the BHT-RL algorithm is equivalent to $\pi^{\TN{MDP}}$. Practically, for someone utilizing the algorithm, the choice of $P(H_0)$ depend on how likely they think that the environment is that of a bandit, based on domain knowledge.
Then given we have run $k$ episodes already we can compute the posterior probabilities for the hypotheses, $P(H_0 | \MC{H}_{t_k} )$ and $P(H_1 | \MC{H}_{t_k} )$. 
\begin{equation*}
	P(H_0 | \MC{H}_{t_k} ) = \frac{ P(H_0, \MC{H}_T) }{ P( \MC{H}_{t_k} ) }
\end{equation*}
\begin{equation*}
	= \frac{ P( \MC{H}_{t_k}| H_0 ) P(H_0) }{ P( \HH_{t_k} | H_0 ) P(H_0) + P( \MC{H}_{t_k} | H_1 ) P(H_1) }
\end{equation*}
The term $\frac{ P( \mathcal{H}_{t_k} | H_1 ) }{ P( \MC{H}_{t_k} | H_0 ) }$ above is the Bayes factor. See Appendix \ref{app:dirichlet} for how we compute the Bayes factor for Dirichlet transition priors. 

\begin{algorithm}
\SetAlgoLined
\SetKwInput{KwInput}{Input}
\KwInput{CB algorithm $\pi^{\TN{CB}}$; 
MDP algorithm $\pi^{\TN{MDP}}$; 
Prior probability $P( H_0 )$; Generative models for state transitions under $H_0$ and $H_1$ respectively.}
 \For{episodes $k=0,1,2,...$}{
  Sample indicator of generative model 
  $B_k \sim \TN{Bernoulli} \big( P( H_0 | \MC{H}_{t_k} ) \big)$ \\
  \uIf{$B_k = 1$}{
  	Let $\pi_k = \pi^{\TN{CB}}_k$ \\ 
   }
  \uElse{
  	Let $\pi_k = \pi^{\TN{MDP}}_k$ \\ 
  }
  \For{timesteps $h=1,2,...,H$}{
   Sample and apply action $A_t \sim \pi_k(S_{t_k+h}, h)$ \\
   Observe $R_{t_k+h}$ and $S_{t_k+h+1}$
  }
  Update both $\pi_k^{\TN{CB}}$ and $\pi_k^{\TN{MDP}}$ with data $\{ S_{t_k+h}, A_{t_k+h}, R_{t_k+h} \}_{h=1}^H$ observed in the episode.
 }
 \caption{Bayesian Hypothesis Testing Reinforcement Learning (BHT-RL)}
 \label{alg:bhtrl}
\end{algorithm}

Note that Bayesian Hypothesis testing approach can be used with any choice of contextual bandit algorithm and MDP based algorithm; the generative model for transition probabilities is only used to compute posterior probability $P( H_0 | \HH_{t_k} )$.
If one chooses posterior sampling methods for the CB and MDP algorithms, then BHT-RL can be interpreted as posterior sampling with a hierarchical prior. Under posterior sampling, a prior is put on the parameters of the environment $\theta$.
The policy for that episode is selected by first sampling $\tilde{\theta} \sim Q( \cdot | \MC{H}_{t_k} )$, where $Q( \cdot | \MC{H}_{t_k} )$ is the posterior distribution over $\theta$. Then the policy for the episode $\pi_k$ is chosen to be the optimal policy for environment $\tilde{\theta}$. When using BHT-RL with posterior sampling CB and MDP algorithms, we have that $\pi_k^{\TN{CB}}$ is the optimal policy for $\tilde{\theta} \sim Q( \cdot | \MC{H}_{t_k}, H_0 )$, the posterior distribution of $Q$ given that the null hypothesis $H_0$ is true. Similarly, $\pi_k^{\TN{MDP}}$ is the optimal policy for $\tilde{\theta} \sim Q( \cdot | \MC{H}_{t_k}, H_1 )$.

%

\subsection{Leveraging endogenous and exogenous features}
\label{sec:endo_exo_decomposition}

In many sequential decision making problems, we may know with certainty that the transition probabilities for some state variables do not change depending on the choice of action. This separation into endogenous and exogenous features allows us to learn more efficiently by only testing the null hypothesis for the features which may potentially be affected by actions. For example, in a mobile health setting, we may be uncertain as to whether or not sending a message affects future user engagement, but we do not believe that our messages can affect the future weather. Formally, we assume that the state space $\mathcal{S}$ can be decomposed into two parts: $\mathcal{S} = \mathcal{X} \times \mathcal{Z}$, where $\mathcal{X}$ includes all known exogenous variables and $\mathcal{Z}$ includes all potentially endogenous variables. We assume that $S_t = [X_t, Z_t]$ for $X_t \in \MC{X}$ and $Z_t \in \MC{Z}$ and that for all $S_t, S_{t+1}$ and $A_t$,
\begin{equation*}
	P_\theta(S_{t+1}  | S_t, A_t) = P_\theta(X_{t+1}, Z_{t+1} | X_t, Z_t, A_t) 
\end{equation*}
\begin{equation*}
	 = P_\theta(X_{t+1}|X_t) P_\theta(Z_{t+1} | Z_t, A_t)
\end{equation*}
For the reward model, we assume that $E_\theta[ R_t | S_t, A_t ]$, the expected reward in a given state and action, is affected by both $X_t$ and $Z_t$. Note that our assumptions differ from other recent works which decompose the state space into endogenous and exogenous components, because we still assume that the exogenous states $X_t$ can affect the value of the expected reward \citep{dietterich2018exogenous,chitnis2020learning}.

Our BHT-RL approach takes advantage of this decomposition to endogenous vs. exogenous feature decomposition by only performing Bayesian hypothesis testing regarding the transition probabilities of sub-states $Z_t$, which are potentially endogenous. Performing Bayesian hypothesis testing only on the subset of potentially endogenous states leads to more efficient learning of whether the environment is an CB vs. MDP, compared to performing hypothesis testing regarding the transition probabilities for the entire state $S_t$. Moreover, this decomposition can allow BHT-RL to easily scale to much large state spaces when a large number of state features are already known to be exogenous.

\subsection{Regret Guarantees}

We now define regret in the episodic setting. 
We first define the value function, which is the expected value of following some policy $\pi$ during an episode:
\begin{equation*}
	V_{\pi, h}^\theta( s ) = \E_{\pi} \bigg[ \sum_{h'=h}^H R ( S_{h'}, A_{h'} ) \bigg| \theta \bigg]
\end{equation*}
We define $\pi^*(\theta)$ to be the optimal policy for some MDP (or contextual bandit) environment $\theta$; barring computational issues, the optimal policy for a given MDP environment can be solved for using dynamic programming.

The frequentist regret is defined as the difference in total expected reward for the optimal policy versus the actual policy used:
\begin{equation*}
	\MC{R}_K( \pi, \theta ) = \sum_{k=0}^K \sum_{s \in \MC{S}} \rho(s) \big( V_{\pi^*, 1}^\theta (s) - V_{\pi_k, 1}^\theta (s) \big)
\end{equation*}
Above $\rho(s)$ represents the probability of starting the episode in state $s$, so $\sum_{s \in \MC{S}} \rho(s)$.
For the Bayesian regret, we assume the MDP environment $\theta$ is drawn from prior distribution $Q$. The Bayesian regret is defined as follows:
\begin{equation*}
	\MC{BR}_K( \pi, \theta ) = \E_{\theta \sim Q} [ \MC{R}_K( \pi, \theta ) ]
\end{equation*}
Note, frequentist regret bounds are automatically Bayesian regret bounds, as they must hold for the worst case environment $\theta$.
Bayesian regret bounds generally assume that the algorithm knows the prior on the environment $Q$. 
\begin{theorem}[Bayesian Regret Bound for MDP Posterior Sampling]
	\label{thm:psrlregret}
	Let $Q$ be the prior distribution over $\theta$ used by the MDP posterior sampling algorithm. Let rewards $R_t \in [0, C]$, for some constant $0 < C < \infty$. Then for $T = (K+1) H$,
	\begin{equation*}
		\MC{BR}_K( \pi, \theta ) = \E_{\theta \sim Q} [ \MC{R}_K( \pi, \theta ) ]
		= O (HS \sqrt{AT \log (SAT) })
	\end{equation*}
\end{theorem}
\citet{osband2013psrl} prove that posterior sampling on MDPs has Bayesian regret $\tilde{O} (HS \sqrt{AT})$, as stated in Theorem \ref{thm:psrlregret} above. Since BHT-RL with posterior sampling contextual bandit and MDP algorithms is simply posterior sampling with a hierarchical prior, we can apply the regret bound of Theorem \ref{thm:psrlregret}. Thus, BHT-RL with posterior sampling CB and MDP algorithms has Bayesian regret $\tilde{O} (HS \sqrt{AT})$, as stated in Corollary \ref{cor:bhtregret}. In other words, Corollary \ref{cor:bhtregret} below follows directly from Theorem \ref{thm:psrlregret} because BHT-RL with CB and MDP algorithms is equivalent to posterior sampling with prior distribution $Q := P(H_0) Q( \cdot | H_0 ) + P(H_1) Q( \cdot | H_1 )$. 
\begin{corollary}[Bayesian Regret Bound for BHT-RL with Posterior Sampling]
	\label{cor:bhtregret}
	Suppose we use BHT-RL with posterior sampling CB and MDP algorithms. Let $P(H_0) \in [0,1]$ be the prior probability of null hypothesis. Let $Q( \cdot | H_0 )$ and $Q( \cdot | H_1 )$ be the prior distribution over $\theta$ conditional on the null and alternative hypotheses respectively.  When rewards $R_t \in [0, C]$, for some constant $0 < C < \infty$, 
	\begin{equation*}
		\MC{BR}_m( \pi, \theta ) = \E_{\theta \sim Q} [ \MC{R}_m( \pi, \theta ) ]
		= O (HS \sqrt{AT \log (SAT) })
	\end{equation*}
	where distribution $Q$ over $\theta$ is defined as $Q := P(H_0) Q( \cdot | H_0 ) + P(H_1) Q( \cdot | H_1 )$.
\end{corollary}

\begin{figure}[b]
  	\centerline{ 
  	\includegraphics[width=\linewidth]{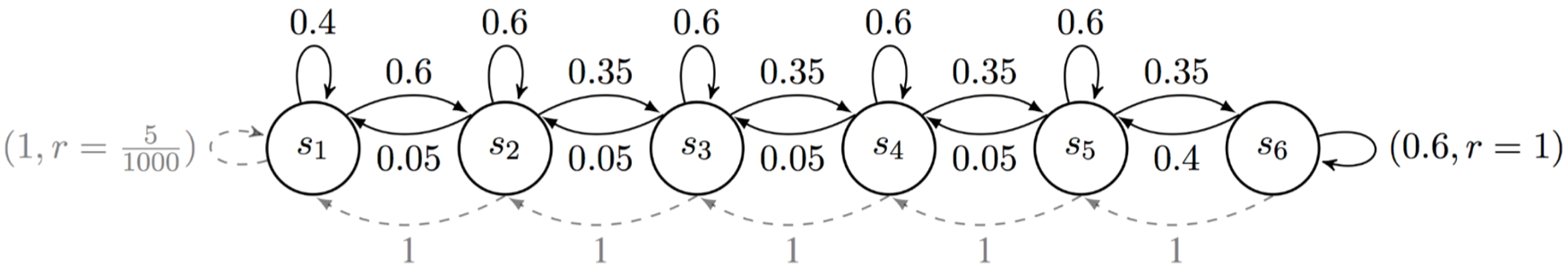} }
  	\caption{\bo{River Swim.} Figure from \citet{osband2013psrl}.}
  	\label{fig:riverswimenv}
\end{figure}

\section{Experiments}


\begin{figure*}[h]
  	\centerline{ 
  		\includegraphics[width=0.25\linewidth]{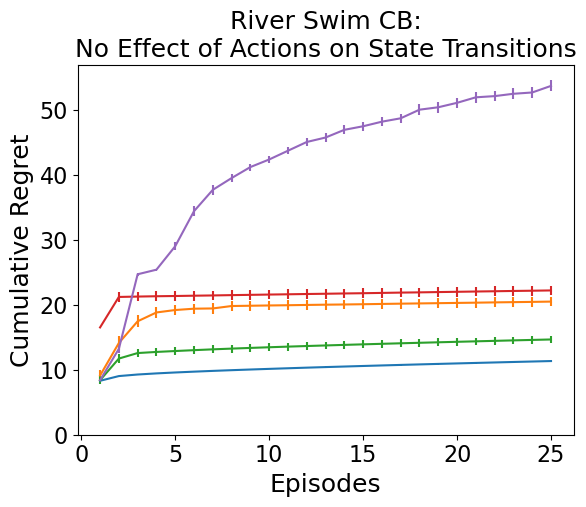} 
		~ \includegraphics[width=0.28\linewidth]{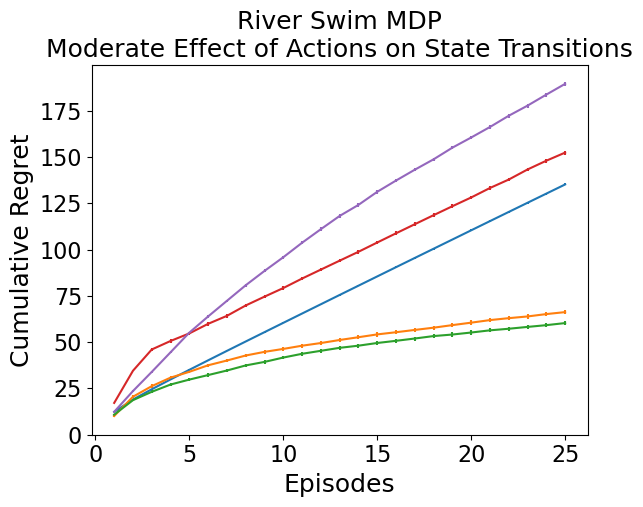} 
		\includegraphics[width=0.445\linewidth]{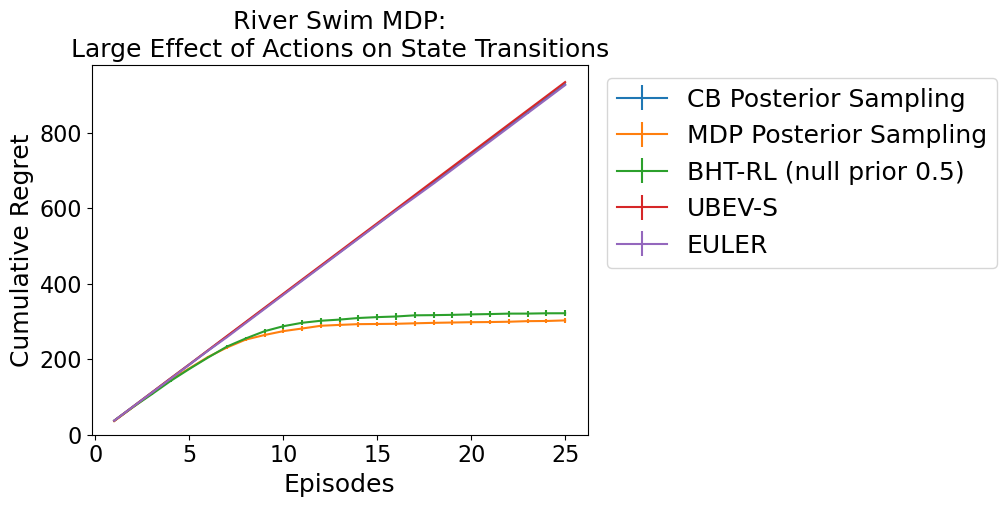}
	}
  	\centerline{ 
  		\includegraphics[width=0.255\linewidth]{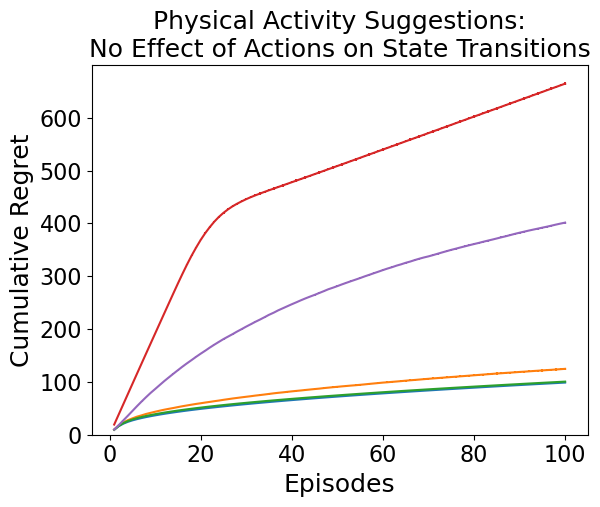} 
		~ \includegraphics[width=0.275\linewidth]{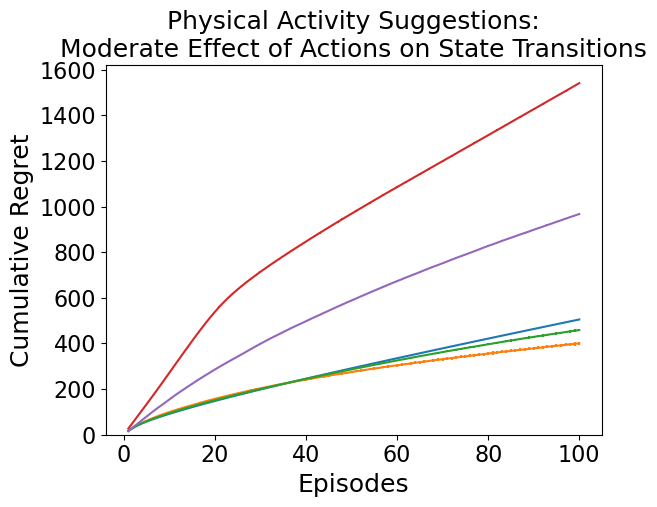}
		\includegraphics[width=0.44\linewidth]{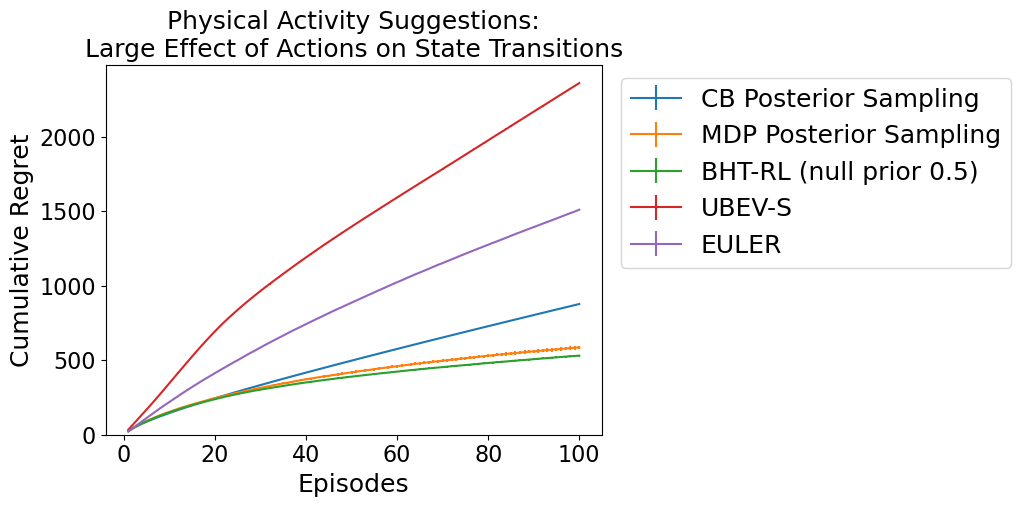}
	}
  	\caption{\bo{Cumulative regret of BHT-RL compared with baselines.} The cumulative regret of the different algorithms is plotted vs. the number of episodes for the riverswim (top) and physical activity suggestions (bottom) environments. For each environment we plot the performance of its CB (left), MDP (right) and intermediate (center) variants. For the posterior sampling algorithms, the Dirichlet prior $\bs{\alpha}$ vector is set to all $1$s. In all cases, BHT-RL's performance is comparable with the best performing algorithm. Error bars denote standard errors over $100$ repetitions.}
  	\label{fig:regret_vs_episodes}
\end{figure*}

We run experiments in simulation environments to demonstrate the advantages of using BHT-RL when the nature of the environment is unknown. On the environments---a toy riverswim domain, randomly generated CBs / MDPs (see Appendix \ref{app:simulations}), and a model inspired by a real world mobile-health application---we show that BHT-RL significantly outperforms CB algorithms in MDP environments and vice versa, while performing nearly as well as CB algorithms in a bandit setting (and similarly for MDPs). Furthermore, we show our methods compare favorably with upper-bound based state of the arts methods aimed at adressing the same problem.

\subsection{Environments}

\paragraph{Riverswim.} The first MDP environment we consider is the river swim environment introduced in \citet{osband2013psrl}, and illustrated in Figure~\ref{fig:riverswimenv}. The optimal policy in this environment must take a series of optimal actions to reach the high reward state on the right, and therefore a bandit algorithm which does not consider the long term benefits of actions will perform very poorly in it. In order to compare with a similar bandit environment, we construct a ``CB River Swim environment'' in which the the transition probability between any two states is uniform and independent of the action, while the rewards for each state are equivalent to those of the original MDP. 

To test the performance of different algorithms as a function of their ``banditness'', we interpolate between the two environments by constructing domains with the following transition function:
\begin{align}
\label{eq:cb_to_mdp_spectrum}
    P_{\lambda} = (1 - \lambda)P_{\text{CB}} + \lambda P_{\text{MDP}},
\end{align}
where $P_{\text{CB}}$ and $P_{\text{MDP}}$ are the transition functions for the CB and MDP environments respectively. Thus, $\lambda = 1$ reduces to the original MDP environment, and as $\lambda \rightarrow 0$, the environment resembles more and more a CB, as the effect of actions on the transition probabilities diminishes.

\paragraph{Mobile Health Physical Activity Suggestions Environment.}

We consider a more realistic simulation environment, which is motivated by the mobile health problem of learning when to send activity suggestions to users. Highly sedentary lifestyles are associated with increased rates of many diseases including cardiovascular disease and diabetes \citep{hu2003sedentary,biswas2015sedentary}. Health apps, through notifications delivered to mobile phones, smart watches and other wearable devices, have increasingly been used to remind users to take walks in order to encourage physical activity. RL is particularly important for learning when to send activity suggestions to users because of the high rate at  which users stop using mobile health applications \citep{eysenbach2005law}. The RL algorithm must learn to send messages when users will be receptive to activity suggestions and not send messages when users will find messages bothersome.

In this simulation setting, several times a day the RL algorithm must decide whether or not to send the user a message encouraging them to take a walk. The algorithm must learn in which contexts to send messages in order to maximize the physical activity of the user. The contextual information we include is detailed in Table \ref{tab:statefeatures}. The reward is the log-step count in a fixed time period following the decision time. Our choice of features and reward is inspired by real world mobile health implementations, particularly \cite{liao2020personalized}, who recently ran a mobile health study encouraging physical activity among people with hypertension. Additionally, the step count goal feature is inspired by the FitBit, which by default includes hourly step goals for users. 

\begin{table}[h]
\begin{center}
    \begin{tabular}{|c|c|c|c|}
     \hline
     \bo{Context Variable} & \bo{Values} \\
     \hline
     Time of Day & Morning, Afternoon, Evening \\ \hline
     Weather & Fair, Poor \\ \hline
     Engagement & Engaged, Disengaged \\ \hline
     Reached Step Goal & Goal Met, Goal Missed \\ \hline
    \end{tabular}
    \caption{Values contextual variables can take ($24$ total states).}
    \label{tab:statefeatures}
\end{center}
\end{table}

In our simulation environment, the expected reward for sending a message is generally greater than or equal to the expected reward for not sending a message. However, while sending a message generally increases the immediate reward, it may increase the probability that the user transitions to a low reward state. In particular, if messages are sent when users are disengaged, this leads to a higher probability of transitioning to a disengaged state. However, if messages are sent when users are engaged, this leads to high reward and a high probability of remaining engaged. We use this model for the transition probabilities to reflect how users can both become more engaged over time when messages are sent at opportune times and how users can become more annoyed with messages if sent at inconvenient times. 
We construct two base environments---a CB and an MDP, and adjust the environment by modifying how much actions can affect state transition probabilities using the same method as in Equation~\eqref{eq:cb_to_mdp_spectrum}, e.g., how much previous actions can affect users' future probability of being engaged with the app and reaching their step goals; see Appendix \ref{app:simulations}.

In this environment we make use of the exogenous vs. endogenous decomposition of state describes in Section~\ref{sec:endo_exo_decomposition}. Specifically, we treat time of day and weather as exogenous variables which are known to be unaffected by the agent's actions, and are therefore not included in the Bayesian hypothesis testing component.



\begin{figure}[t]
  	\centerline{ 
  	    \includegraphics[width=0.48\linewidth]{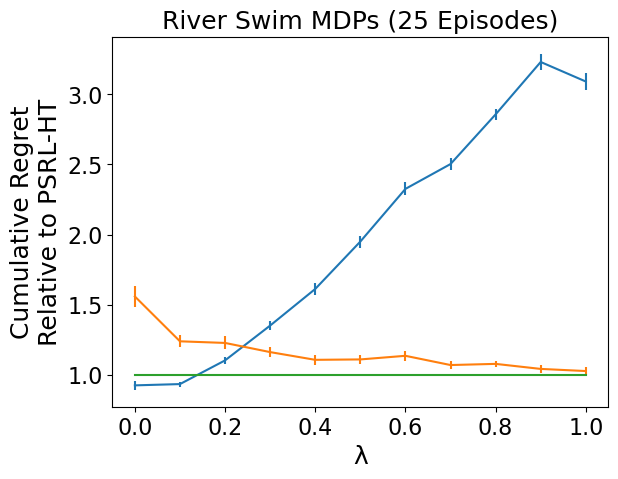}  
  	    \includegraphics[width=0.52\linewidth]{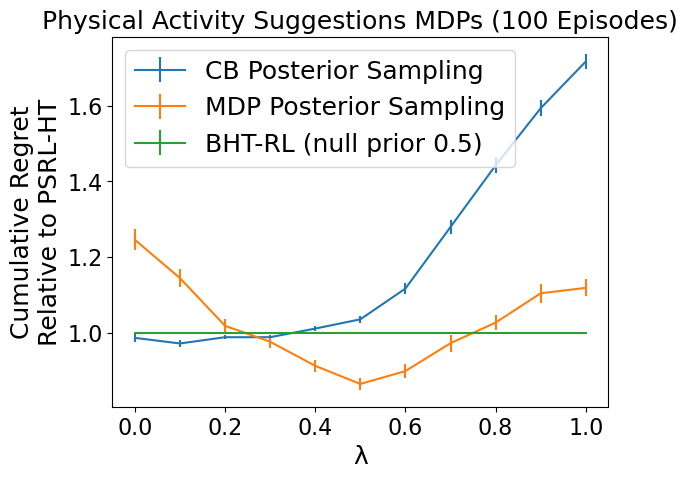}
	}		
  	\caption{\bo{Performance as a function of ``banditness''.}
  	Cumulative regret at $H=100$ relative to BHT-RL in the riverswim (left) and physical activity (right) environments vs. the interpolation parameter $\lambda$ with Dirichlet prior $\bs{\alpha}=1$. Error bars denote standard errors over $100$ repetitions.}
  	\label{fig:regret_vs_lambda}
\end{figure}


\begin{figure}[h]
  	\centerline{ 
  	    \includegraphics[width=0.45\linewidth]{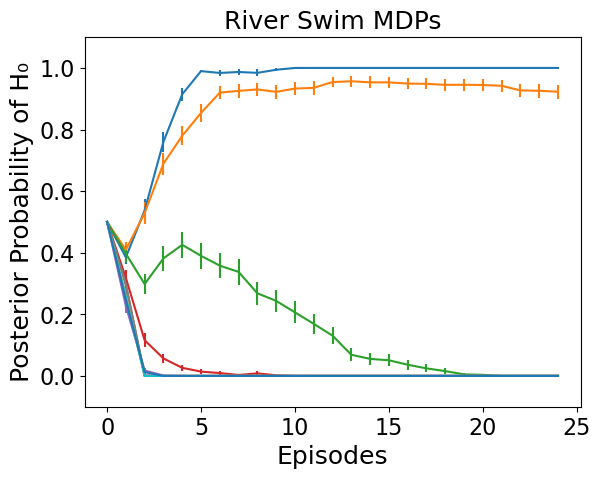}
  	    \includegraphics[width=0.55\linewidth]{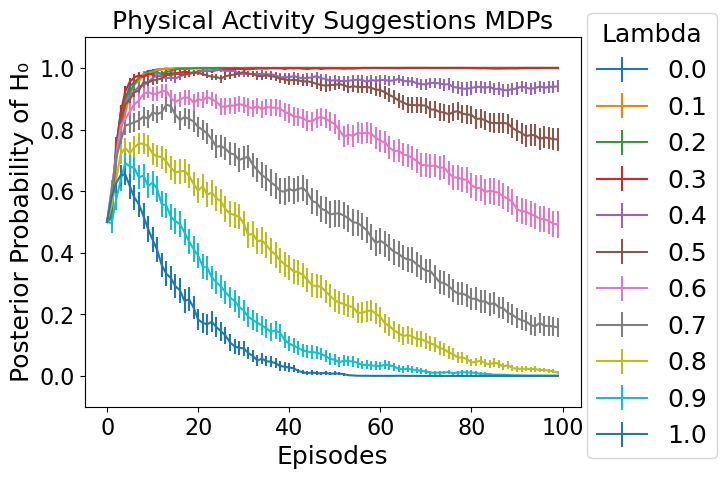} 
	}		
  	\caption{\bo{Posterior probability of the null hypothesis.} The posterior probability of the null hypothesis in the riverswim (left) and physical activity (right) environments evolves with the number of episodes for different interpolation parameters $\lambda$. We set $H=100$ and Dirichlet prior $\bs{\alpha}=1$. Error bars denote standard errors over $100$ repetitions.}
  	\label{fig:null_proba_vs_ep}
\end{figure}


\begin{figure}[t]
  	\centerline{ 
		\includegraphics[width=0.5\linewidth]{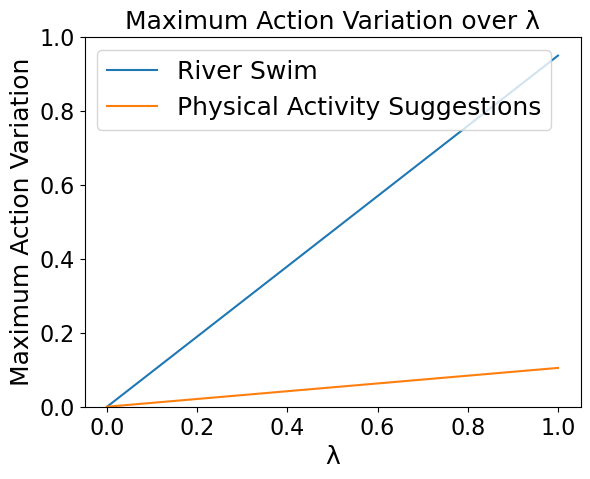}
	}
  	\caption{\bo{Maximum action variation for different environments over $\lambda$.} }
  	\label{fig:action_variation}
\end{figure}


\begin{figure*}[h]
  	\centerline{ 
  		\includegraphics[width=0.29\linewidth]{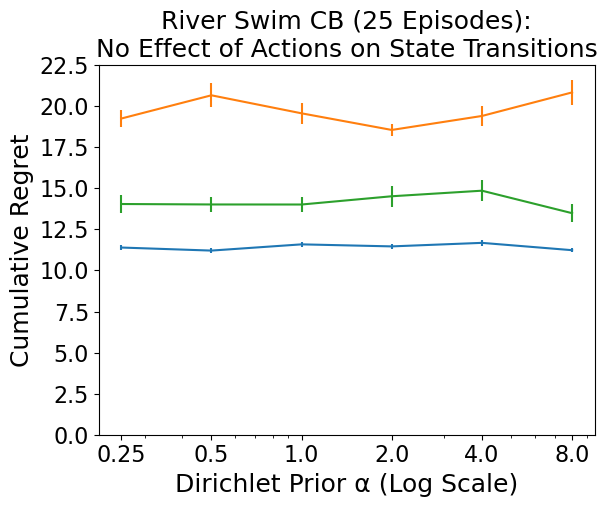} 
		~ \includegraphics[width=0.3\linewidth]{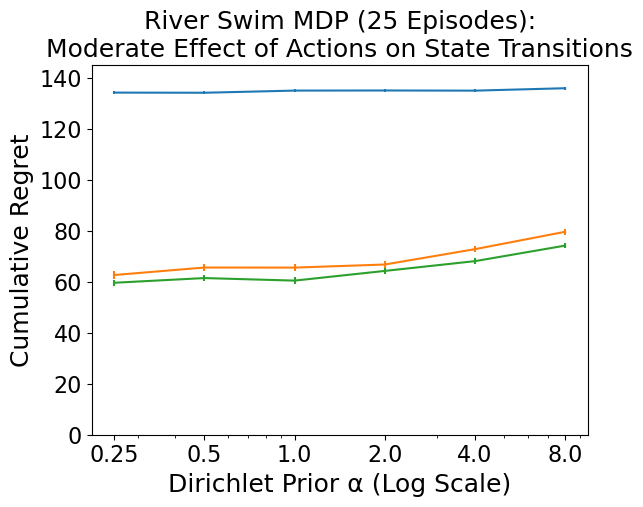} 
		\includegraphics[width=0.3\linewidth]{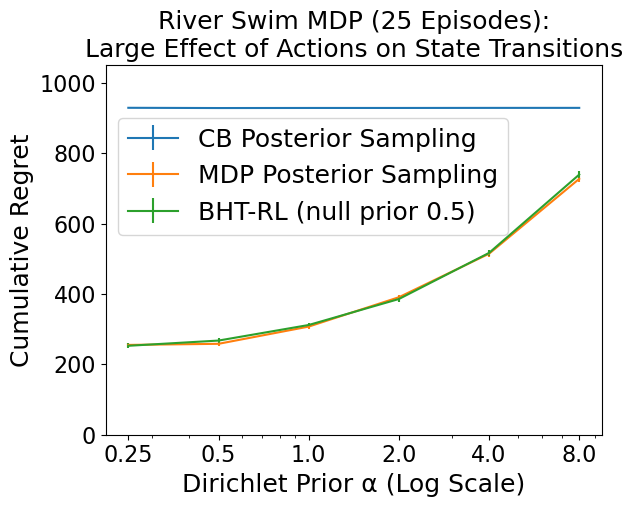}
	}		
  	\centerline{ 
  		\includegraphics[width=0.3\linewidth]{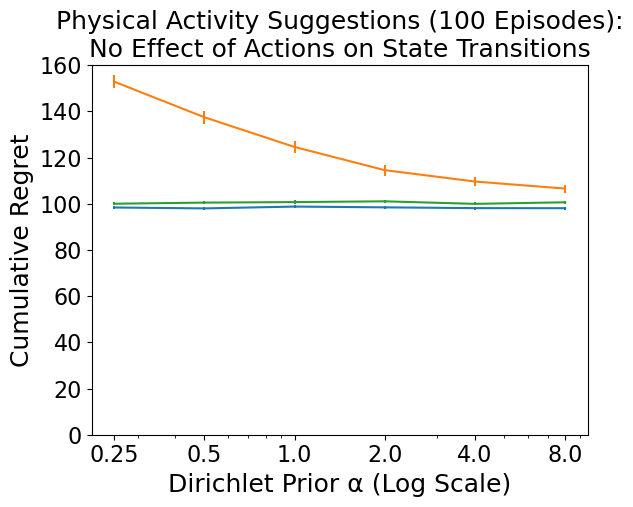} 
		~~~ \includegraphics[width=0.3\linewidth]{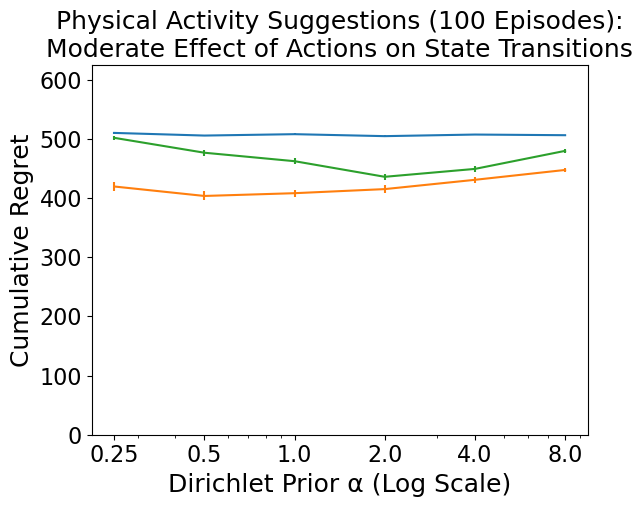} 
		\includegraphics[width=0.3\linewidth]{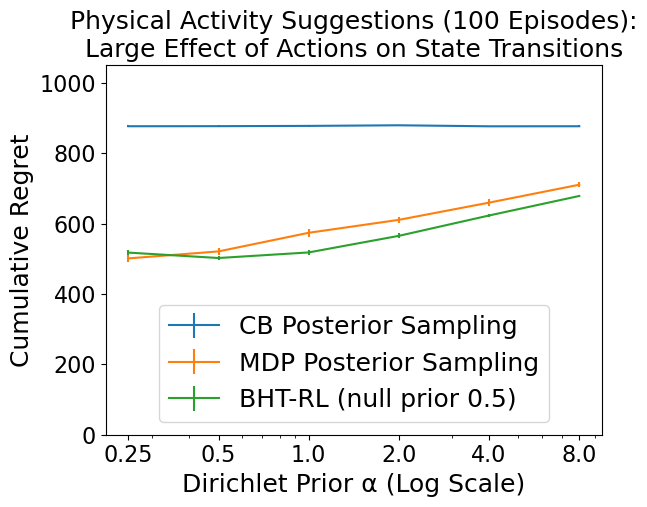}
	}		
  	\caption{\bo{Sensitivity to algorithm hyper-parameters.}
  	We plot the cumulative regret at $H=100$ as a function of the Dirichlet hyper-parameter $\bs{\alpha}$ for the riverswim (top) and physical activity suggestions (bottom) environments. For each environment we plot the results for its CB (left), MDP (right) and intermediate (center) variants.}
  	\label{fig:regret_vs_alpha}
\end{figure*}

\subsection{Results}

\paragraph{BHT-RL consistently outperforms CB and MDP algorithms in their respective misspecified environments.} 
In Figure~\ref{fig:regret_vs_episodes} we plot the cumulative regret over episodes for both environments. For each environments three variants are used---the pure CB and pure MDP variants, as well as an intermediate environment in which the impact of actions on the transition probabilities is smaller than for the original MDP variant of the environment ($\lambda =0.6$). 
As noted in Figure~\ref{fig:cb_vs_mdp}, the performance of the CB and MDP posterior sampling (PS) algorithms is optimal when the algorithm is used for the correct environment (lower regret is better). However, we see that in all cases BHT-RL outperforms the CB and MDP posterior sampling algorithms in their respective misspecified environments, i.e., $\lambda$ close to $0$ for MDP-PS and $\lambda$ close to $1$ for CB-PS. Moreover, in many cases the BHT-RL performs comparable to, if not better, than the algorithm explicitly designed for the particular environment. 
This demonstrates the ability of BHT-RL to perform well without knowledge of the nature of the environment. 
Furthermore, comparison with the UBEV-S \citep{zanette2018structure} and EULER \citep{zanette2019tighter} baselines, which are also designed to operate when the nature of the environment is unknown, shows that they under-perform because of their reliance on confidence bounds which may be very loose.

We note that while the distinction between a bandit environment and an MDP is well defined, we can blur the definition and ask how ``bandit-like'' an MDP is by considering how strong of an effect do actions have on transition probabilities. In this sense, the $\lambda$ parameter in equation~\eqref{eq:cb_to_mdp_spectrum} controls how close to a bandit our environment is, where the examples shown in Figure~\ref{fig:cb_vs_mdp} correspond to $\lambda=0$ and $1$. 
To better quantify the performance of the different posterior sampling methods as a function of the ``banditness'' of the environment, in Figure \ref{fig:regret_vs_lambda} we plot the cumulative regret after $H=100$ episodes as a function of $\lambda$. For clarity of presentation, we plot the ratio between the regret for the CB or MDP algorithms over the regret of BHT-RL. 
BHT-RL consistently performs better than CB-PS and MDP-PS in their corresponding misspecified environments. For $\lambda$ close to $1$, BHT-RL outperforms CB-PS because through the Bayesian hypothesis testing procedure BHT-RL is able to learn that the environment is an MDP environment, while CB-PS is not able to ever learn the optimal policy. For $\lambda$ close to $0$, BHT-RL outperforms MDP-PS because the Bayesian hypothesis testing procedure regularizes the BHT-RL algorithm by reducing the planning horizon, which leads it to incur lower regret. BHT-RL consistently outperforms the worst performing algorithm (out of CB-PS and MDP-PS) across all values of $\lambda$. Moreover, BHT-RL has better or comparable performance to the best performing algorithm for most values of $\lambda$, with the exception of intermediate values of $\lambda$ in the physical activity suggestion environment which we will be better equipped to discuss following the next section.





\paragraph{Using BHT-RL to learn about the nature of the environment.}

An additional benefit of BHT-RL is that it naturally outputs a posterior estimate for the probability of the null hypothesis that actions do not affect state transitions. In Figure~\ref{fig:null_proba_vs_ep} we plot the posterior probability of the null hypothesis as a function of the number of episodes. This knowledge can be useful in practice if we would like to consider what additional algorithms or methods to apply to the domain, and is obtained at no additional computational cost. This is in line with the Bayesian hypothesis testing literature (in contrast to frequentist hypothesis testing), which uses cutoff values for the Bayes factor to reject the null hypothesis \citep{quintana2018bayesian}.


As a way to illustrate the relative difficulties of distinguishing between CB and MDP in the two simulation environments, following the terminology used in \citet{jiang2016structural}, we define the maximum action variation for an environment as the following:
\begin{equation*}
    \max_{a, a' \in \MC{A}; s, s' \in \MC{S}} 
    \big| P_\theta( s' | s, a) - P_\theta( s' | s, a') \big|
\end{equation*}
Note that for small non-zero values of $\lambda$, which correspond to smaller maximum action variation values, the posterior probability of $H_0$ decays to zero very slowly, indicating that a large amount of data is needed to rule out $H_0$ if the effect of actions on transition probabilities is small.
As shown in Figure \ref{fig:action_variation}, the physical activity suggestions environment has a much smaller maximum action variation compared to river swim. The difficulty of learning whether the environment is a CB vs. MDP in the physical activity suggestions environment, is further demonstrated in the plots of the posterior probability of null hypothesis in Figure \ref{fig:null_proba_vs_ep}. We believe that the increased difficulty of learning whether the environment is a CB vs. MDP is what leads MDP-PS to outperform BHT-RL for intermediate values of $\lambda$ in Figure \ref{fig:regret_vs_lambda}.

\paragraph{Sensitivity to algorithm hyper-parameters.}

We demonstrate the sensitivity of our algorithm to $\alpha$. the parameter determining the Dirichlet distribution of the prior over transitions for the MDP algorithm. In Figure~\ref{fig:regret_vs_alpha} we plot the regret of the different PS algorithms for the different environments. As expected, the prior on transition probabilities does not effect the performance in bandit settings, but may affect performance in an MDP environment where being able to model the transition probabilities well is important.

While the value of $\alpha$ affects the performance of both the MDP-PS and BHT-RL algorithms, BHT-RL consistently performs better than or comparable to (1) CB algorithms in MDP environments and (2) MDP algorithms in CB environments for all values of $\alpha$. BHT-RL also performs at least as well as the best performing algorithm in each environment with the exception is for very small values of $\alpha$, where MDP-PS has lower cumulative regret. 
We believe this is due to the fact that a very small value of $\alpha$ pushes the MDP-PS and the MDP component of BHT-RL to learn very sparse transition probabilities, which are inconsistent with the true environment. This in turn leads the BHT-RL algorithm to require more examples to learn whether the environment is an MDP or CB, and therefore incurs greater regret; see Figure \ref{fig:pH0_vs_alpha} in the Appendix. Note that this can be mitigated by choosing a larger value of $\alpha$.


\section{Discussion}
Our simulation results show that at least in finite state MDP and contextual bandit environments, the BHT-RL algorithm can perform well even when it is unknown whether the environment is that of an MDP or contextual bandit. Additionally, the BHT-RL approach allows practitioners to easily incorporate prior knowledge about the environment dynamics into their algorithm. 
Finally, since BHT-RL stochastically reduces the planning horizon, it can also be used as a regularization method for the full MDP based algorithm.

Some limitations of our work are that our method assumes the stationarity of the dynamics of the RL environment. Thus, our method is not robust to non-stationarity, which is often encountered in real world sequential decision making problems. 
To adapt the method to a continuous state setting would additionally require a model of the transition probabilities, which may be unrealistic to assume is known for real world problems. Additional open questions are whether it is possible to show additional theoretical guarantees regarding the BHT-RL algorithm, like a frequentist regret bound or a regret bound when the prior is misspecified.


Beyond just learning whether the environment is that of a contextual bandit or an MDP, we conjecture that bayesian hypothesis testing could also be used to address other aspects of reinforcement learning problems.
One example is learning better state representations \citep{ortner2019regret}, which is a major open problem in the reinforcement learning field \citep{dulac2019challenges}.




\bibliography{mdp_bht_uai}

\clearpage
\appendix

\onecolumn

\section{Bayesian Hypothesis Testing for Dirichlet Priors on Transition Probabilities}
\label{app:dirichlet}
We define the set of states as $\MC{S}$ and the set of actions as $\MC{A}$. 
Suppose we have data $\HH_T = \{ S_t, A_t, R_t \}_{t=1}^T$. 

\begin{itemize}
	\item \bo{Null hypothesis} $H_0$: We model our data as follows 
		\begin{itemize}
			\item For each $s \in \MC{S}$ we draw $\bs{\tilde{\theta}}_s \sim \TN{Dirichlet}( \bs{\alpha} )$
			\item For all $t \in [1 \colon T]$ such that $S_t = s$, we have that $S_{t+1} \sim \TN{Categorical}( \bs{\tilde{\theta}}_s )$
		\end{itemize}
	\item \bo{Alternative hypothesis} $H_1$: We model our data as follows 
		\begin{itemize}
			\item For each $s \in \MC{S}$ and each $a \in \MC{A}$ we draw $\bs{\tilde{\theta}}_{s,a} \sim \TN{Dirichlet}( \bs{\alpha} )$
			\item For all $t \in [1 \colon T]$ such that $S_t = s$ and $A_t = a$, we have that $S_{t+1} \sim \TN{Categorical}( \bs{\tilde{\theta}}_{s,a} )$
		\end{itemize}
\end{itemize}

We choose prior probabilities over the hypotheses $P(H_0)$ and $P(H_1) = 1 - P(H_0)$. 
Then we can calculate the posterior probabilities $P(H_0 | \MC{H}_T )$ and $P(H_1 | \MC{H}_T )$
\begin{equation*}
	P(H_0 | \HH_T ) = \frac{ P(H_0, \HH_T) }{ P( \HH_T ) }
	= \frac{ P( \HH_T| H_0 ) P(H_0) }{ P( \HH_T | H_0 ) P(H_0) + P( \HH_T | H_1 ) P(H_1) }
	= \frac{ 1 }{ 1 + K }
\end{equation*}
where $K =  \frac{ P( \HH_T | H_1 ) P(H_1) }{ P( \HH_T | H_0 ) P(H_0) }$ is the Bayes factor. \\

Let us now derive the posterior distributions. Let $\bs{\theta}$ represent all transition probability parameters, so $\theta = \{ \bs{\theta}_s \}_{s \in \MC{S}} \cup \{ \bs{\theta}_{s,a} \}_{s \in \MC{S}, a \in \MC{A}}$. 
\begin{equation*}
	P( \bs{\theta} | \HH_T ) = \frac{ P ( \bs{\theta}, \HH_T ) }{ P ( \HH_T ) } 
	= \frac{ P( \HH_T | \bs{\theta} ) P (\bs{\theta}) }{ \int P( \HH_T | \bs{\theta} ) P( \bs{\theta} ) d \bs{\theta} } 
	=: \frac{X}{Y}
\end{equation*}
First examining the numerator term $X$,
\begin{equation*}
	X = P( \HH_T | \bs{\theta} ) \big[ P (\bs{\theta} | H_0) P( H_0) + P (\bs{\theta} | H_1) P( H_1) \big]
\end{equation*} 
\begin{multline*}
	= P( H_0) \prod_{s \in \MC{S}} \bigg[ \TN{Dirichlet} ( \bs{\theta}_s ; \bs{\alpha} ) \prod_{t=1}^T \TN{Categorical}(S_{t+1}; \bs{\theta}_s )^{ \1_{ S_t = s } } \bigg] \\
	+ P( H_1 ) \prod_{s \in \MC{S}} \prod_{a \in \MC{A}} \bigg[ \TN{Dirichlet} ( \bs{\theta}_{s,a} ; \bs{\alpha} ) \prod_{t=1}^T \TN{Categorical}(S_{t+1}; \bs{\theta}_{s,a} )^{\1_{S_t = s, A_t = a}} \bigg]
\end{multline*}
Below $B(\bs{\alpha})$ is the multivariate beta function.
\begin{multline*}
	= \frac{ P( H_0 ) }{ B( \bs{\alpha} )^{|\MC{S}|} } \prod_{s \in \MC{S}} \bigg[ \prod_{s' \in \MC{S}} \theta_s(s')^{\alpha(s') - 1} 
		\prod_{t=1}^T \theta_s(s')^{ \1_{ S_t = s, S_{t+1} = s'} } \bigg] \\
	+ \frac{ P( H_1 ) }{ B( \bs{\alpha} )^{|\MC{S}| |\MC{A}|}  } \prod_{s \in \MC{S}} \prod_{a \in \MC{A}} \bigg[ \prod_{s' \in \MC{S}} \theta_{s,a}(s')^{\alpha(s') - 1} 
		\prod_{t=1}^T \theta_{s,a}(s')^{ \1_{S_t = s, A_t = a, S_{t+1}=s'} } \bigg]
\end{multline*}
\begin{equation*}
	= \frac{ P( H_0 ) }{ B( \bs{\alpha} )^{|\MC{S}|} } \prod_{s \in \MC{S}} \bigg[ \prod_{s' \in \MC{S}} \theta_s(s')^{\alpha(s') - 1 + \sum_{t=1}^T \1_{ S_t = s, S_{t+1} = s'} } 
		 \bigg] 
	+ \frac{ P( H_1 ) }{ B( \bs{\alpha} )^{|\MC{S}| |\MC{A}|}  } \prod_{s \in \MC{S}} \prod_{a \in \MC{A}} \bigg[ \prod_{s' \in \MC{S}} \theta_{s,a}(s')^{\alpha(s') - 1 + \sum_{t=1}^T \1_{S_t = s, A_t = a, S_{t+1}=s'} } \bigg]
\end{equation*}
\begin{equation*}
	= \frac{ P( H_0 ) }{ B( \bs{\alpha} )^{|\MC{S}|} } \prod_{s \in \MC{S}} \bigg[ \prod_{s' \in \MC{S}} \theta_s(s')^{\alpha(s') - 1 + \sum_{t=1}^T \1_{ S_t = s, S_{t+1} = s'} } 
		 \bigg] \\
	+ \frac{ P( H_1 ) }{ B( \bs{\alpha} )^{|\MC{S}| |\MC{A}|}  } \prod_{s \in \MC{S}} \prod_{a \in \MC{A}} \bigg[ \prod_{s' \in \MC{S}} \theta_{s,a}(s')^{\alpha(s') - 1 + \sum_{t=1}^T \1_{S_t = s, A_t = a, S_{t+1}=s'} } \bigg]
\end{equation*}
We define $\bo{N}_s = [ \sum_{t=1}^T \1_{S_t = s, S_{t+1}=1}, \sum_{t=1}^T \1_{S_t = s, S_{t+1}=2}, ..., \sum_{t=1}^T \1_{S_t = s, S_{t+1}=|\MC{S}|}]$ and \\
$\bo{N}_{s,a} = [ \sum_{t=1}^T \1_{S_t = s, A_t=a, S_{t+1}=1}, \sum_{t=1}^T \1_{S_t = s, A_t=a, S_{t+1}=2}, ..., \sum_{t=1}^T \1_{S_t = s, A_t=a, S_{t+1}=|\MC{S}|}]$.
\begin{equation*}
	= \frac{ P( H_0 ) }{ B( \bs{\alpha} )^{|\MC{S}|} } \prod_{s \in \MC{S}} B \big( \bs{\alpha} + \bo{N}_s \big) \TN{Dirichlet} \big( \bs{\theta}_s; \bs{\alpha} + \bo{N}_s \big) 
	+ \frac{ P( H_1 ) }{ B( \bs{\alpha} )^{|\MC{S}| |\MC{A}|}  } \prod_{s \in \MC{S}} \prod_{a \in \MC{A}} B \big( \bs{\alpha} + \bo{N}_{s,a} \big) \TN{Dirichlet} \big( \bs{\theta}_{s,a}; \bs{\alpha} + \bo{N}_{s,a} \big)
\end{equation*}
Thus,
\begin{multline*}
	X = \frac{ P( H_0 ) B( \bs{\alpha} )^{|\MC{S}| ( |\MC{A}| - 1 )} \prod_{s \in \MC{S}} B \big( \bs{\alpha} + \bo{N}_s \big) \TN{Dirichlet} \big( \bs{\theta}_s; \bs{\alpha} + \bo{N}_s \big) }{ B( \bs{\alpha} )^{|\MC{S}| |\MC{A}|} } \\
		+ \frac{ P( H_1 ) \prod_{s \in \MC{S}} \prod_{a \in \MC{A}} B \big( \bs{\alpha} + \bo{N}_{s,a} \big) \TN{Dirichlet} \big( \bs{\theta}_{s,a}; \bs{\alpha} + \bo{N}_{s,a} \big) }{ B( \bs{\alpha} )^{|\MC{S}| |\MC{A}|} }
\end{multline*}

Since $X = P( \HH_T | \bs{\theta} ) P (\bs{\theta})$ and $Y= \int P( \HH_T | \bs{\theta} ) P( \bs{\theta} ) d \bs{\theta}$, we have that
\begin{equation*}
	Y =  \frac{ P( H_0 ) }{ B( \bs{\alpha} )^{|\MC{S}|} } \prod_{s \in \MC{S}} B \big( \bs{\alpha} + \bo{N}_s \big)
	+ \frac{ P( H_1 ) }{ B( \bs{\alpha} )^{|\MC{S}| |\MC{A}|}  } \prod_{s \in \MC{S}} \prod_{a \in \MC{A}} B \big( \bs{\alpha} + \bo{N}_{s,a} \big)
\end{equation*} 
\begin{equation*}
	= \frac{ P( H_0 ) B( \bs{\alpha} )^{|\MC{S}| ( |\MC{A}| - 1 )} \prod_{s \in \MC{S}} B \big( \bs{\alpha} + \bo{N}_s \big) 
		+  P( H_1 ) \prod_{s \in \MC{S}} \prod_{a \in \MC{A}} B \big( \bs{\alpha} + \bo{N}_{s,a} \big) }{ B( \bs{\alpha} )^{|\MC{S}| |\MC{A}|} }
	=: \frac{ W_0 + W_1 }{ B( \bs{\alpha} )^{|\MC{S}| |\MC{A}|} }
\end{equation*} 

Thus, 
\begin{equation*}
	P( \bs{\theta} | \HH_T ) = \frac{X}{Y} 
	= \frac{W_0}{W_0 + W_1} \prod_{s \in \MC{S}} \TN{Dirichlet} \big( \bs{\theta}_s; \bs{\alpha} + \bo{N}_s \big) 
		+ \frac{W_0}{W_0 + W_1} \prod_{s \in \MC{S}} \prod_{a \in \MC{A}} \TN{Dirichlet} \big( \bs{\theta}_{s,a}; \bs{\alpha} + \bo{N}_{s,a} \big)
\end{equation*} 

Note that 
\begin{equation*}
	P( H_0 | \HH_T ) = \frac{ P(H_0 | \HH_T) P(\HH_T )}{P(\HH_T )} 
	= \frac{ P( \HH_T | H_0 ) P( H_0 ) }{ P(\HH_T ) } 
	= \frac{W_0}{W_0 + W_1}
\end{equation*} 
\begin{equation*}
	P( H_1 | \HH_T) = \frac{W_1}{W_0 + W_1}
\end{equation*}

\onecolumn

\section{Simulation Details}
\label{app:simulations}
\subsection{Riverswim Environments}

\begin{itemize}
	\item We add $\N(0,0.01)$ noise to all rewards.
	\item $P_{\TN{MDP}}$ transitions are those of the original river swim environment as in Figure \ref{fig:riverswimenv}.
	\item $P_{\TN{CB}}$ transitions are uniform over all states, i.e, $P( S_{t+1}=s | S_{t}=s', A_t=a ) = \frac{1}{|\MC{S}|}$ for all $s, s' \in \MC{S}$ and $a \in \MC{A}$.
\end{itemize}

\paragraph{Algorithm Hyper-Parameters}
\begin{itemize}
	\item For Bandit and MDP posterior sampling we have independent $\N(1, 1)$ priors on the rewards.
	\item For MDP posterior sampling we have Dirichlet ($\bs{\alpha} \in \real^S$) priors on the transition probabilities.
	\item For BHT-PSRL we set the probability of the null hypothesis to $P(H_0) = 0.5$.
	\item UBEV-S and EULER we choose failure probability $\delta = 0.1$.
\end{itemize}

\subsection{Physical Activity Suggestions Environments}

\begin{itemize}
    \item Reward is log step count
	\item Actions are binary: $0$ means no message sent; $1$ means message sent
	\item We add $\N(0,0.01)$ noise on rewards
\end{itemize}

\begin{table}[h]
\begin{center}
    \begin{tabular}{|c|c|c|c|}
     \hline
     \bo{Context Variable} & \bo{Values} & \bo{Variable Notation} & \bo{Endogenous vs. Exogenous} \\
     \hline
     Time of Day & Morning ($0$), Afternoon ($1$), Evening ($2$) & $S_{t}^{\TN{time}}$ & Exogenous \\ \hline
     Weather & Fair ($0$), Poor ($1$) & $S_{t}^{\TN{weather}}$ & Exogenous \\ \hline
     Engagement & Disengaged ($0$), Engaged ($1$) & $S_{t}^{\TN{engagement}}$ & Endogenous \\ \hline
     Reached Step Goal & Goal Missed ($0$), Goal Met ($1$) & $S_{t}^{\TN{goal}}$ & Endogenous \\ \hline
    \end{tabular}
    \caption{Values that contextual variables can take. There are $24$ distinct states in total.}
\end{center}
\end{table}

\paragraph{Endogenous vs. Exogenous State Variables}
\begin{itemize}
	\item In our simulation environment, state space $\mathcal{S}$ can be decomposed into two parts: $\mathcal{S} = \mathcal{X} \times \mathcal{Z}$ such that $\MC{X}$ state variables are exogenous and $\MC{Z}$ state variables are potentially endogenous in the following sense:
    \begin{equation*}
        P(S_{t+1} | S_t, A_t) = P(X_{t+1}, Z_{t+1} | X_t, Z_t, A_t) 
        = P(X_{t+1}|X_t) P(Z_{t+1} | Z_t, A_t)
    \end{equation*}
	\item For the purposes of Bayesian hypothesis PSRL approach, the bayesian hypothesis testing only has to occur for substates $Z_t$, rather than the full state $S_t$, which improves the performance and stability of our approach.
	\item For the model of the reward, we assume that $E[ R_t | S_t, A_t ]$, the expected reward in a given state and action, is affected by both $X_t$ and $Z_t$. For example, how much a user walks following a decision time can be affected by both endogenous and exogenous variables.
\end{itemize}

\paragraph{State Transition Probabilities}

\begin{table}[b]
\begin{center}
    \begin{tabular}{|c|c|c|c|}
     \hline
      & \bo{Morning} & \bo{Afternoon} & \bo{Evening} \\
     \hline
     \bo{Morning} & 0 & 1 & 0 \\ \hline
     \bo{Afternoon} & 0 & 0 & 1 \\ \hline
     \bo{Evening} & 1 & 0 & 0 \\ \hline
    \end{tabular}
    \caption{Time of Day Transition Probability Matrix}
\end{center}
\end{table}

\clearpage
\begin{table}[h]
\begin{center}
    \begin{tabular}{|c|c|c|c|}
     \hline
      & \bo{Fair Weather} & \bo{Poor Weather} \\
     \hline
     \bo{Fair Weather} & 0.6 & 0.4 \\ \hline
     \bo{Poor Weather} & 0.3 & 0.7 \\ \hline
    \end{tabular}
    \caption{Weather Transition Probability Matrix}
\end{center}
\end{table}
\begin{table}[h]
\begin{center}
    \begin{tabular}{|c|c|c|c|c|}
     \hline
      & \bo{Disengaged,} & \bo{Disengaged,} & \bo{Engaged,} & \bo{Engaged,} \\
      & \bo{Goal Missed} & \bo{Goal Met} & \bo{Goal Missed} & \bo{Goal Met} \\
     \hline
     \bo{Disengaged, Goal Missed} & 0.35 & 0.35 & 0.15 & 0.15 \\ \hline
     \bo{Disengaged, Goal Met} & 0.4 & 0.25 & 0.2 & 0.15 \\ \hline
     \bo{Engaged, Goal Missed} & 0.2 & 0.25 & 0.3 & 0.25 \\ \hline
     \bo{Engaged, Goal Met} & 0.15 & 0.15 & 0.3 & 0.4 \\ \hline
    \end{tabular}
    \caption{Endogenous Variable Transition Probability Matrix under $A_t = 1$, i.e., $P( Z_{t+1} | Z_t, A_t=1)$}
\end{center}
\end{table}
\begin{table}[h]
\begin{center}
    \begin{tabular}{|c|c|c|c|c|}
     \hline
      & \bo{Disengaged,} & \bo{Disengaged,} & \bo{Engaged,} & \bo{Engaged,} \\
      & \bo{Goal Missed} & \bo{Goal Met} & \bo{Goal Missed} & \bo{Goal Met} \\
     \hline
     \bo{Disengaged, Goal Missed} & 0.45 & 0.35 & 0.1 & 0.1 \\ \hline
     \bo{Disengaged, Goal Met} & 0.5 & 0.3 & 0.15 & 0.05 \\ \hline
     \bo{Engaged, Goal Missed} & 0.05 & 0.3 & 0.3 & 0.35 \\ \hline
     \bo{Engaged, Goal Met} & 0.05 & 0.05 & 0.35 & 0.55 \\ \hline
    \end{tabular}
    \caption{Endogenous Variable Transition Probability Matrix under $A_t = 0$, i.e., $P( Z_{t+1} | Z_t, A_t=0)$}
\end{center}
\end{table}

\begin{itemize}
    \item Above we state the transition probabilities under $P_{\TN{MDP}}$
    \item For transition probabilities $P_{\TN{CB}}$, we simply set $P( Z_{t+1} | Z_t, A_t=1)$ to the corresponding values of $P( Z_{t+1} | Z_t, A_t=0)$ under $P_{\TN{MDP}}$.
\end{itemize}

\paragraph{Expected Reward}
\begin{equation*}
    E[R_t | S_t, A_t=a] = 
    \sum_{i=0}^2 \theta_{a,i}^{\TN{time}} \1_{i=S_{t}^{\TN{time}}}
    + \sum_{j=0}^1 \theta_{a,j}^{\TN{weather}} \1_{j=S_{t}^{\TN{weather}}}
    + \sum_{k=0}^1 \sum_{l=0}^1 \theta_{a,k,l}^{\TN{endogenous}} \1_{k=S_{t}^{\TN{engagement}}} \1_{l=S_{t}^{\TN{goal}}}
\end{equation*}

\begin{itemize}
    \item Time of Day
        \begin{itemize}
            \item $\theta_{0,0}^{\TN{time}} = 0.001$;~ $\theta_{0,1}^{\TN{time}} = 0.01$;~ $\theta_{0,2}^{\TN{time}} = 0.005$
            \item $\theta_{1,0}^{\TN{time}} = 0.001$;~ $\theta_{1,1}^{\TN{time}} = 0.02$;~ $\theta_{1,2}^{\TN{time}} = 0.01$
        \end{itemize}
    \item Weather
        \begin{itemize}
            \item $\theta_{0,0}^{\TN{weather}} = 0.01$;~ $\theta_{0,1}^{\TN{weather}} = 0.015$
            \item $\theta_{1,0}^{\TN{weather}} = 0.01$;~ $\theta_{1,1}^{\TN{weather}} = 0.025$
        \end{itemize}
    \item Endogenous
        \begin{itemize}
            \item $\theta_{0,0,0}^{\TN{endogenous}} = 0.005$;~ $\theta_{0,0,1}^{\TN{endogenous}} = 0.4$;~ 
            $\theta_{0,1,0}^{\TN{endogenous}} = 0.35$;~
            $\theta_{0,1,1}^{\TN{endogenous}} = 2.25$
            \item $\theta_{1,0,0}^{\TN{endogenous}} = 0.01$;~ $\theta_{1,0,1}^{\TN{endogenous}} = 0.405$;~ $\theta_{1,1,0}^{\TN{endogenous}} = 1.75$;~
            $\theta_{1,1,1}^{\TN{endogenous}} = 2.5$
        \end{itemize}
\end{itemize}

\paragraph{Algorithm Hyper-Parameters}
\begin{itemize}
	\item For Bandit and MDP posterior sampling we have independent $\N(1, 1)$ priors on the rewards.
	\item For MDP posterior sampling we have Dirichlet ($\bs{\alpha} \in \real^S$) priors on the transition probabilities.
	\item For BHT-PSRL we set the probability of the null hypothesis to $P(H_0) = 0.5$.
	\item UBEV-S and EULER we choose failure probability $\delta = 0.1$.
\end{itemize}

\subsection{Random MDP Environments}
\begin{itemize}
    \item The following simulation environment is based on that in \citet{jiang2015dependence}.
    \item For these experiments we randomly sampled 100 MDPs with 10 states and 2 actions from a distribution we refer to as Random-MDP, defined as follows. 
    \item $P_{\TN{MDP}}$ transitions are constructed as follows. For each $s \in \MC{S}$ and each $a \in \MC{A}$, the distribution $P(s' | s, a)$ for all $s' \in \MC{S}$ is chosen by selecting 5 non-zero entries uniformly from all 10 states, filling these 5 entries with values sampled from $\TN{Uniform}(0,1)$, and then by normalizing the values to sum to $1$.
    \item $P_{\TN{CB}}$ transitions are constructed by modifying the $P_{\TN{MDP}}$ transition probabilities. In particular, for each $s \in \MC{S}$, the transition probabilities $P(s' | s, a=1)$ is set to $P(s' | s, a=0)$ from the $P_{\TN{MDP}}$ transition probabilities.
    \item Start state is selected uniformly from all $10$ states.
    \item Mean rewards were likewise sampled independently from $\TN{Uniform}(0,1)$.
    \item Rewards have noise $\N(0, 0.01)$.
    \item Results averaged over 100 randomly sampled MDPs.
\end{itemize}

\paragraph{Algorithm Hyper-Parameters}
\begin{itemize}
    \item For Bandit and MDP posterior sampling we have independent $\N(1, 1)$ priors on the rewards.
	\item For MDP posterior sampling we have Dirichlet ($\bs{\alpha} \in \real^S$) priors on the transition probabilities.
	\item For BHT-PSRL we set the probability of the null hypothesis to $P(H_0) = 0.5$.
	\item UBEV-S and EULER we choose failure probability $\delta = 0.1$.
\end{itemize}

\clearpage
\subsection{Additional Simulation Results}

\begin{figure*}[h]
  	\centerline{ 
  		\includegraphics[width=0.31\linewidth]{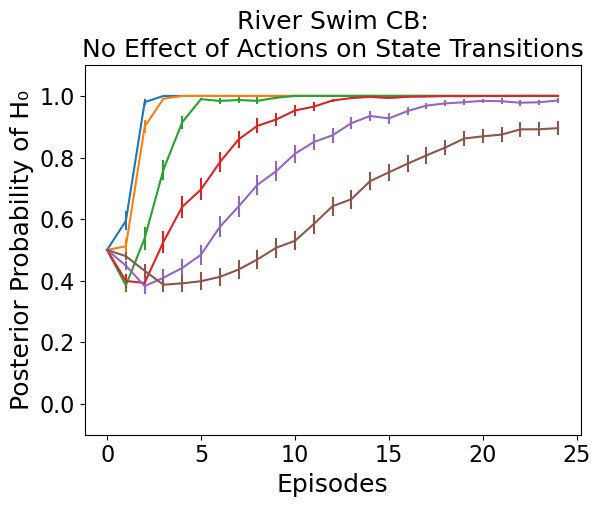} 
		~ \includegraphics[width=0.33\linewidth]{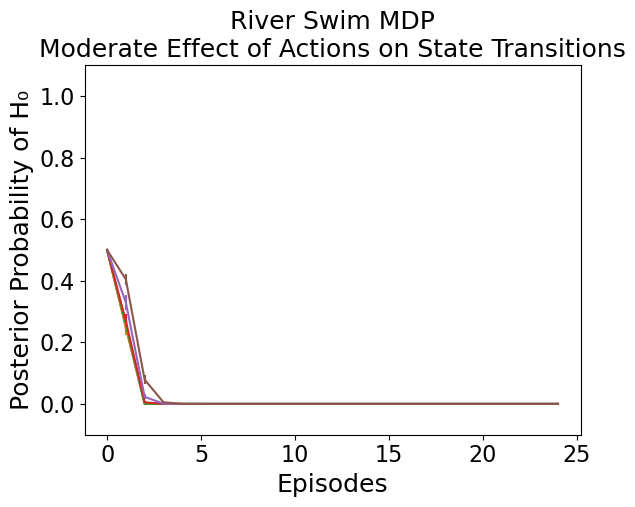} 
		\includegraphics[width=0.33\linewidth]{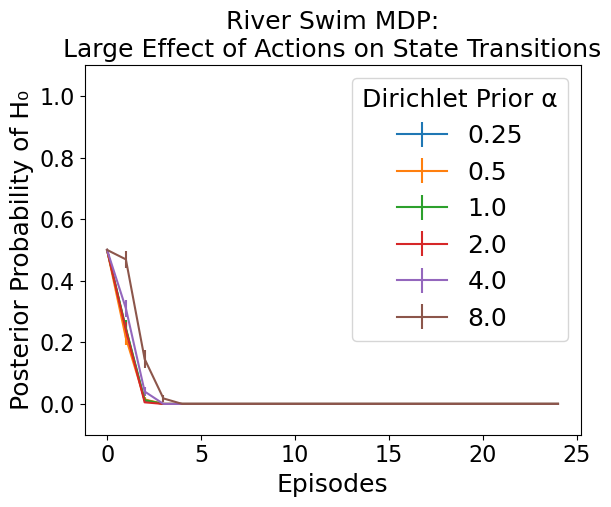}
	}		
  	\centerline{ 
  		\includegraphics[width=0.33\linewidth]{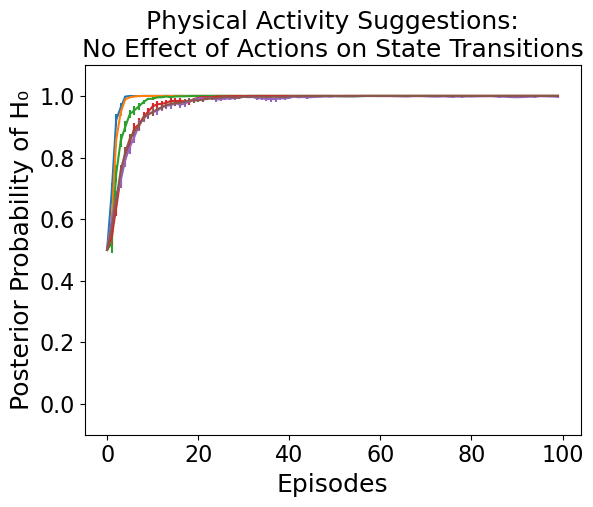} 
		~~~ \includegraphics[width=0.33\linewidth]{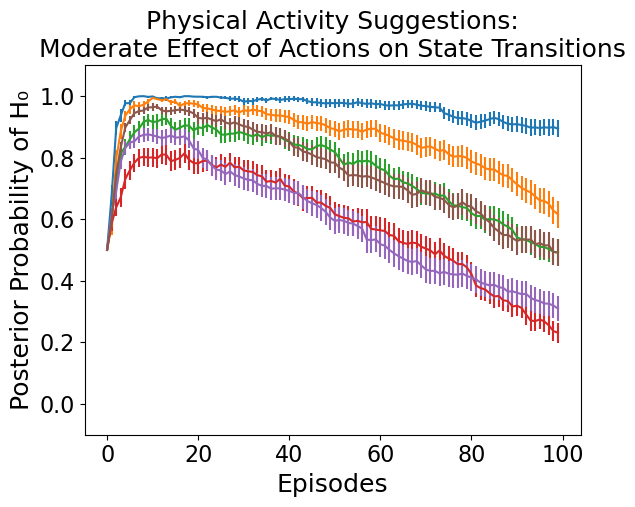} 
		\includegraphics[width=0.33\linewidth]{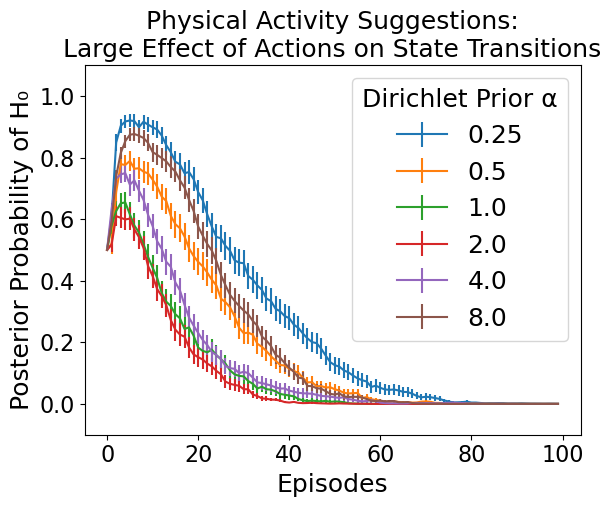}
	}		
	\centerline{ 
  		\includegraphics[width=0.33\linewidth]{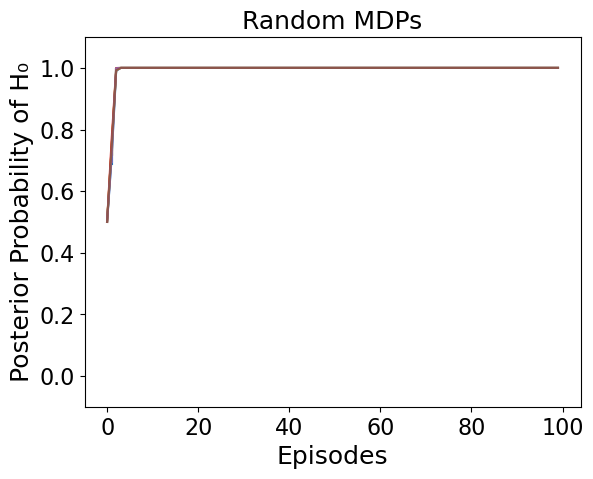} 
		~~~ \includegraphics[width=0.33\linewidth]{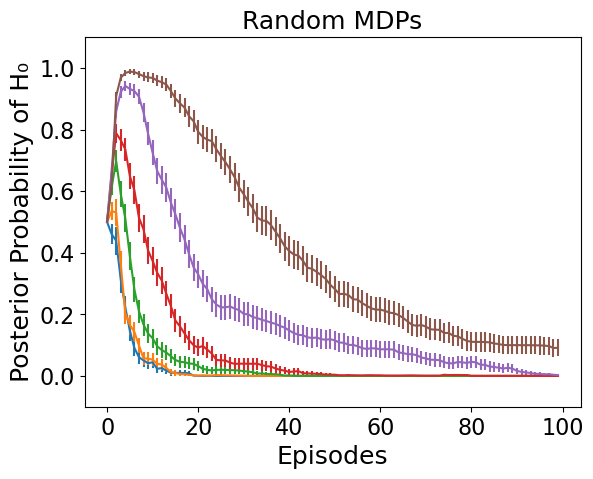} 
		\includegraphics[width=0.33\linewidth]{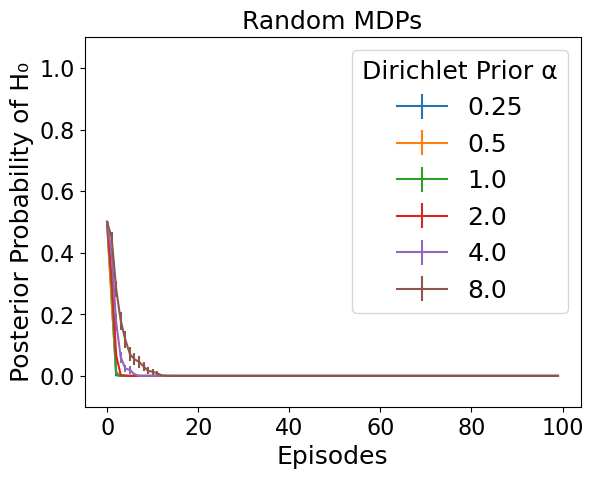}
	}		
  	\caption{\bo{Sensitivity to algorithm hyper-parameters.}
  	We plot how the posterior probability of the null hypothesis for different values of the Dirichlet hyper-parameter $\bs{\alpha}$ for the riverswim (Top), physical activity suggestions (Middle), and random MDP (Bottom) environments. For each environment we plot the results for its CB (Left), MDP (Right) and intermediate (Center) variants.}
  	\label{fig:pH0_vs_alpha}
\end{figure*}



\begin{figure*}[h]
  	\centerline{ 
  		\includegraphics[width=0.26\linewidth]{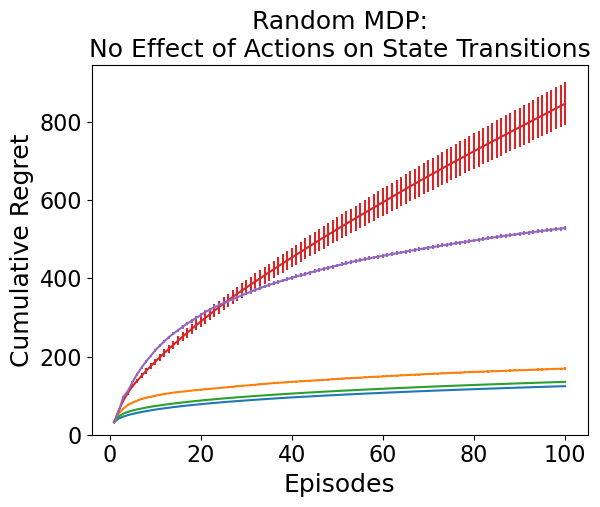} 
		~ \includegraphics[width=0.29\linewidth]{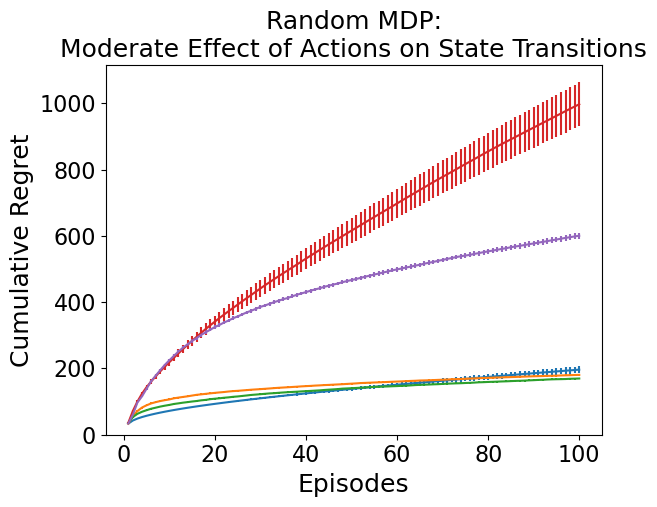} 
		\includegraphics[width=0.455\linewidth]{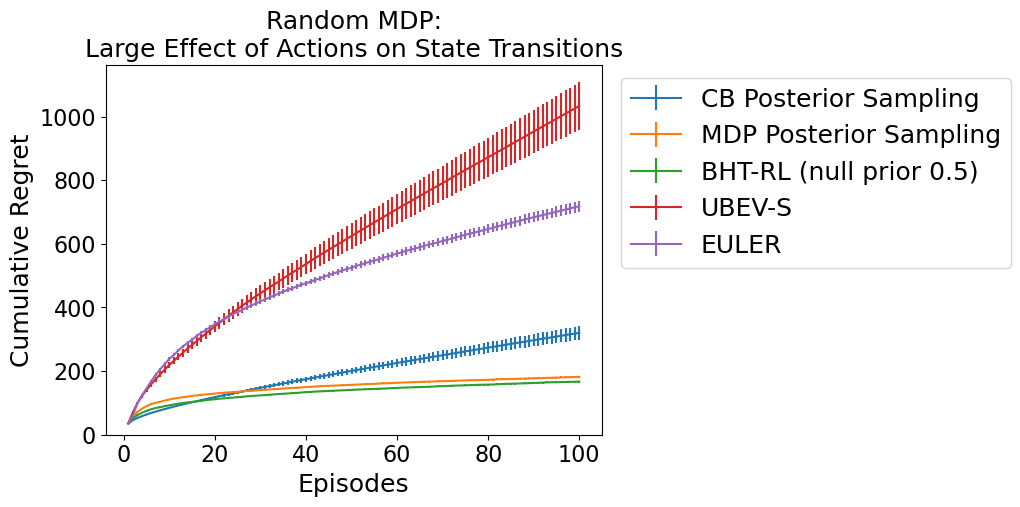}
	}
  	\caption{\bo{Cumulative regret of BHT-RL compared with baselines.} The cumulative regret of the different algorithms is plotted vs. the number of episodes for the riverswim (Top) and physical activity suggestions (Bottom) environments. For each environment we plot the performance of its CB (Left), MDP (Right) and intermediate (Center) variants. For the posterior sampling algorithms, the Dirichlet prior $\bs{\alpha}$ vector is set to all $1$s. In all cases, BHT-RL's performance is comparable with the best performing algorithm.}
  	\label{fig:regret_vs_episodes}
\end{figure*}


\begin{figure}[h]
  	\centerline{ 
  	    \includegraphics[width=0.45\linewidth]{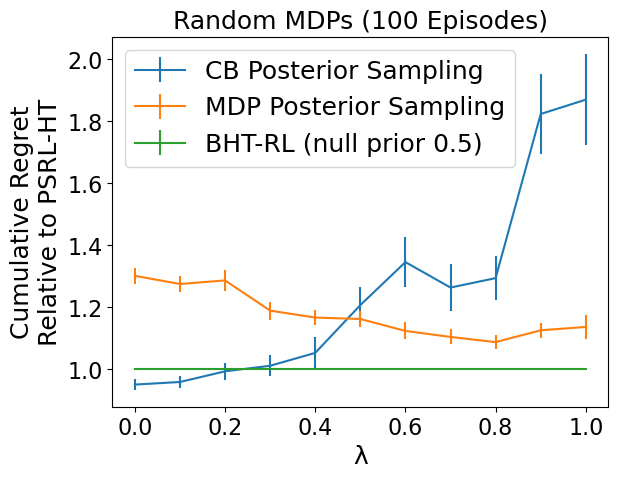}  
	}		
  	\caption{We plot regret relative to PSRL-HT in random MDP environments varying interpolation parameter $\lambda$ with $H=100$. Error bars denote standard errors over $100$ repetitions. Dirichlet prior hyperparameter $\bs{\alpha}=1$.}
  	\label{fig:regret_vs_lambda_random}
\end{figure}


\begin{figure}[h]
  	\centerline{ 
  	    \includegraphics[width=0.48\linewidth]{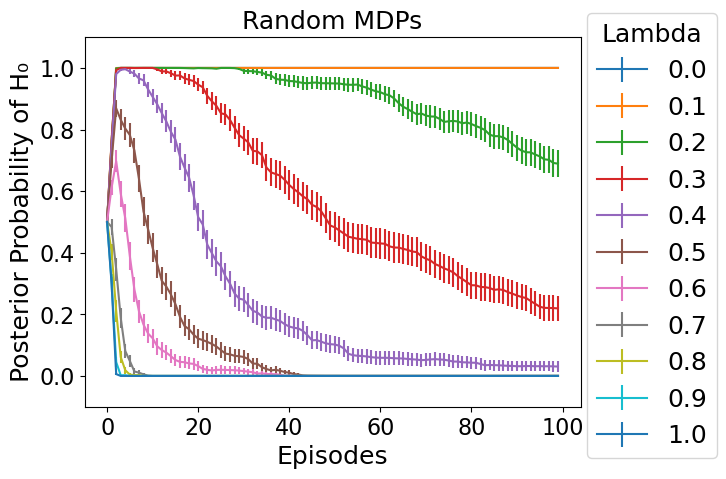}
	}		
  	\caption{We plot posterior probabilities for PSRL-HT in the random MDP environments for different interpolation parameters $\lambda$ with $H=100$. Error bars denote standard errors over $100$ repetitions. Dirichlet prior hyperparameter $\bs{\alpha}=1$.}
  	\label{fig:randomlambda}
\end{figure}


\begin{figure*}[h]
  	\centerline{ 
  		\includegraphics[width=0.31\linewidth]{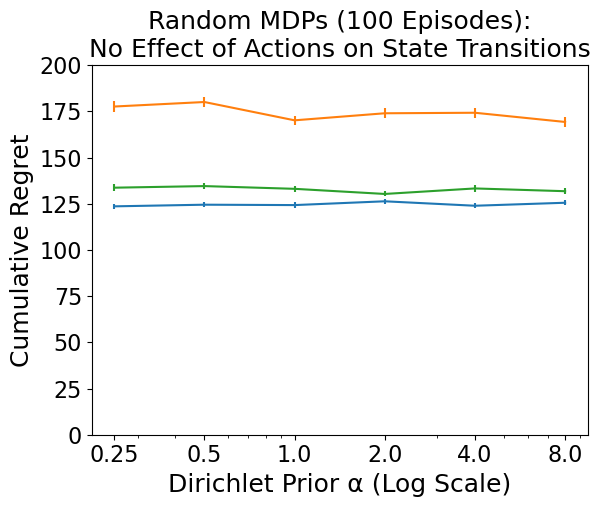} 
		~ \includegraphics[width=0.33\linewidth]{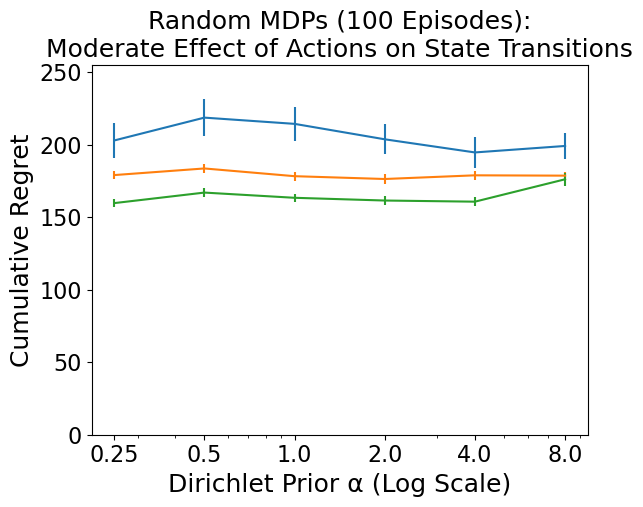} 
		\includegraphics[width=0.33\linewidth]{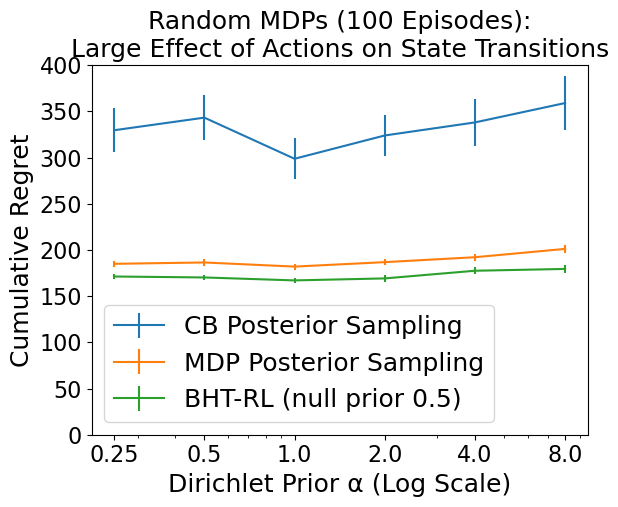}
	}
  	\caption{\bo{Left}: Random CB environments. 
  	\bo{Middle}: transition probabilities interpolated between Random MDP and Random CB environments;
  	\bo{Right}: Random MDP environments. All environments have horizon $H=100$. Error bars denote standard errors over $100$ repetitions. We vary dirichlet prior hyperparameter $\bs{\alpha}$, letting it take values $0.25, 0.5, 1, 2, 4, 8$.}
  	\label{fig:randomalpha}
\end{figure*}


\end{document}